\documentclass[10pt,twocolumn,letterpaper]{article}

 \usepackage[pagenumbers]{cvpr} 

\usepackage[dvipsnames]{xcolor}

\definecolor{cvprblue}{rgb}{0.21,0.49,0.74}
\usepackage[pagebackref,breaklinks,colorlinks,citecolor=cvprblue]{hyperref}
\usepackage{adjustbox}

\def\paperID{*****} 
\def\confName{CVPR}
\def\confYear{2024}

\title{Neural Encoding for Image Recall: Human-Like Memory}

\author{Virgile Foussereau\\
University of California, Berkeley\\
{\tt\small virgile.foussereau@berkeley.edu}
\and
Robin Dumas\\
University of California, Berkeley\\
{\tt\small robin\_dumas@berkeley.edu}
}

\begin{document}
\maketitle
\begin{abstract}
Achieving human-like memory recall in artificial systems remains a challenging frontier in computer vision. Humans demonstrate remarkable ability to recall images after a single exposure, even after being shown thousands of images. However, this capacity diminishes significantly when confronted with non-natural stimuli such as random textures. In this paper, we present a method inspired by human memory processes to bridge this gap between artificial and biological memory systems. Our approach focuses on encoding images to mimic the high-level information retained by the human brain, rather than storing raw pixel data. By adding noise to images before encoding, we introduce variability akin to the non-deterministic nature of human memory encoding. Leveraging pre-trained models' embedding layers, we explore how different architectures encode images and their impact on memory recall. Our method achieves impressive results, with 97\% accuracy on natural images and near-random performance (52\%) on textures. We provide insights into the encoding process and its implications for machine learning memory systems, shedding light on the parallels between human and artificial intelligence memory mechanisms.
\end{abstract}    
\section{Introduction}
\label{sec:intro}

Human memory has been shown to possess a remarkable capacity, capable of storing a massive number of items. Landmark studies in the 1970s demonstrated this prowess by revealing that after viewing 10,000 scenes for a few seconds each, individuals could determine which of two images had been seen with 83\% accuracy \cite{landmark}. \\

\noindent Building upon this foundation, a more recent study performed an experiment where participants were exposed to 2,500 images of real-world objects \cite{oliva}. Subsequently, participants were tested using a two-alternative forced-choice, where they had to distinguish between objects they had seen and novel items. Surprisingly, participants exhibited remarkable memory performance, correctly identifying previously viewed objects with a high accuracy of 93\%. This study shattered the conventional idea that human long-term memory is limited in detail, demonstrating instead the capacity to store a vast number of objects with fidelity. In addition to the forced-choice test, which was the main experiment of the paper as in \cite{landmark}, the researchers also conducted a repeat-detection task. 
In this task, participants were asked to monitor for any repeating images during the presentation of the 2,500 images. This task ensured that participants were actively engaged with the stream of images as they were presented, providing an online estimate of memory storage capacity throughout the entire study session. \\

\noindent This paper aims to achieve human-like memory using neural network encoding. The objective is to find a method that can achieve high accuracy at remembering natural images, but drops to random performance on texture images. We perform a comparison of different methods and encoder models, and provide insights on some failure cases.
\section{Method}
\label{sec:formatting}

\subsection{Latent Space Projection}
\label{sec:latent}

Contrary to traditional computer memory, human memory does not save all the raw pixel values. The memory process involves loss of information and it is mostly the high-level features that are retained. Taking inspiration from this, we base our architecture on a projection of the image in a latent space, through an encoder. Ideally, this encoder should extract the high-level information from the image. We hypothesize that this information should be rich for natural images but low for textures. \\

\noindent We test different pre-trained models as our image encoder: 
\begin{itemize}

\item Contrastive Language-Image Pre-Training (CLIP) is a neural network trained on a large number of (image, text) pairs \cite{CLIP}. Due to its ability to perform a variety of tasks in a zero-shot manner, we believe it is ideally suited to capture information in natural images. Specifically, we use its image encoding module, which is based on a ViT-L/14 Transformer architecture that encodes images in a vector of size 768.

\item AlexNet is a well-known convolutional neural network architecture introduced by Krizhevsky et al. \cite{krizhevsky2012imagenet}. It gained significant attention for its performance in the ImageNet Large Scale Visual Recognition Challenge (ILSVRC) in 2012. Being a much simpler architecture than CLIP, trained on far less data, we include this model as a comparison to get insights on how the encoder choice will affect the performance of our memory system. We use AlexNet to encode each image in a vector of size 1000.

\end{itemize}

\subsection{Memory Perturbation}

\begin{figure}[h]
  \centering
   \includegraphics[width=\linewidth]{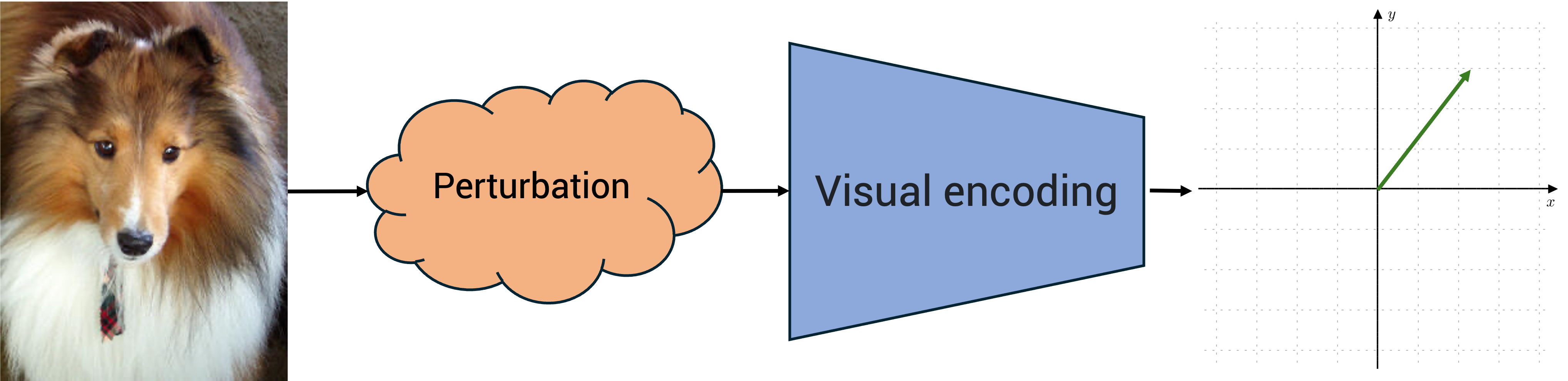}
   \caption{Memory process illustration: perturbation and projection in a latent space. For illustration, the latent space is represented with two dimensions but it is typically much larger.}
   \label{fig:memory_process}
\end{figure}

Using a pre-trained model to encode the image leads to a perfectly deterministic result, meaning that the same image will always give the exact same vector in the latent space. In biology, the memory process is more noisy or blurry. To emulate this effect, we apply a perturbation on the image before its encoding into the memory. This perturbation can either be Gaussian noise of mean zero and standard deviation $\sigma_n$, or Gaussian blur of standard deviation $\sigma_b$. The memory process is illustrated in Fig. \ref{fig:memory_process}. The resulting vectors are stored in a k-d tree \cite{kdtree} that will serve as the system memory.

\subsection{Remembering an image}
\label{ssec:Remembering an image}

At test time, the system is tasked with determining if it has already seen an image or not. The same encoding as in the memory process is used, but without perturbation. Then, a search in the memory k-d tree is performed to get the distance $d_{NN}$ to the nearest neighbor of the vector. This distance $d_{NN}$ will be used depending on the task:

\begin{itemize}
    \item \textbf{Forced-Choice task:} We replicate the main experiment from \cite{oliva}. In this task, our neural system is presented with two images, one previously seen and one novel. The objective is to determine which image was seen before. We calculate the distance $d_{NN}$ for both images, with the image having the smallest distance classified as seen and the other as novel.
    \item \textbf{Repeat-Detection task:} We also replicate the repeat-detection task introduced by \cite{oliva}. Images are presented sequentially in a continuous stream, with repetitions occurring approximately every 1 in 8 images. The neural system's task is to trigger an alarm upon detecting a repetition. We establish a threshold $\delta$ by computing the average $d_{NN}$ for 2,500 seen and 2,500 novel images from a separate dataset. $\delta$ is set at the midpoint between these two values. This method assumes the system has prior exposure to images, similar to human participants in related studies.
\end{itemize}
\section{Results}

\begin{figure}
    \centering
    \includegraphics[width=0.24\linewidth]{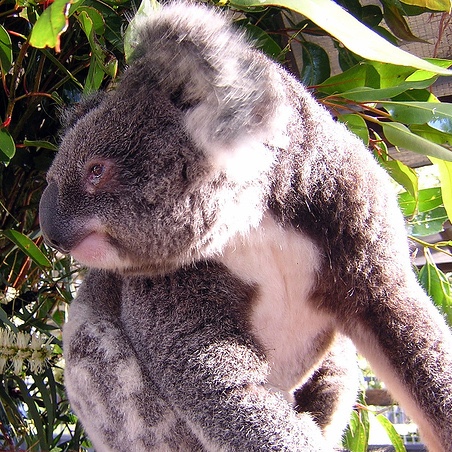}
    \includegraphics[width=0.24\linewidth]{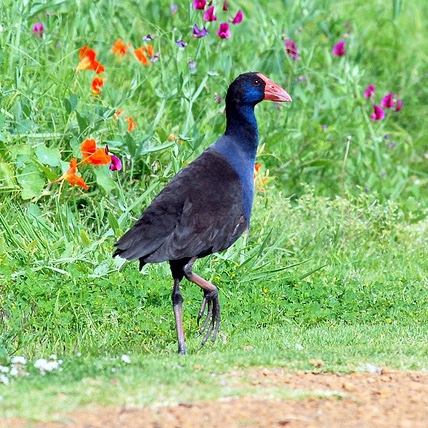}
    \includegraphics[width=0.24\linewidth]{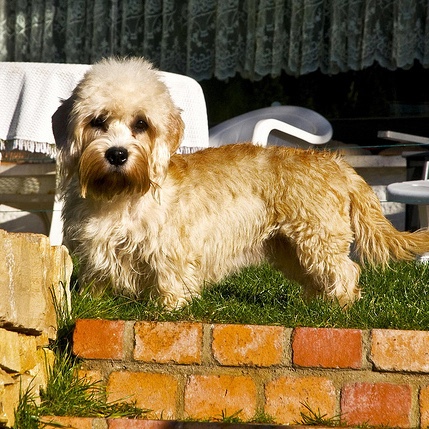}
    \includegraphics[width=0.24\linewidth]{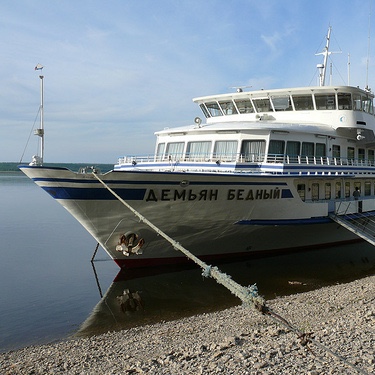}\\
    \includegraphics[width=0.24\linewidth]{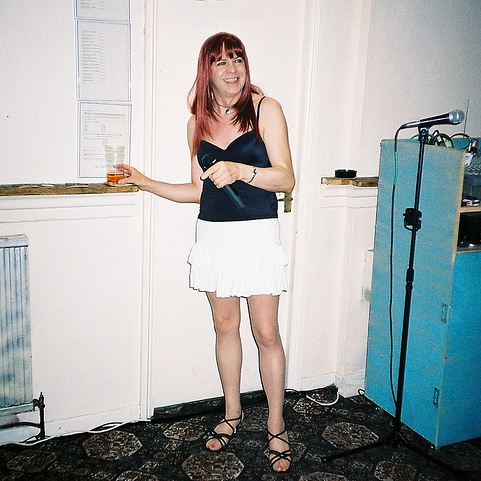}
    \includegraphics[width=0.24\linewidth]{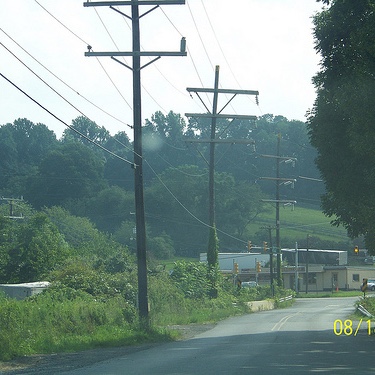}
    \includegraphics[width=0.24\linewidth]{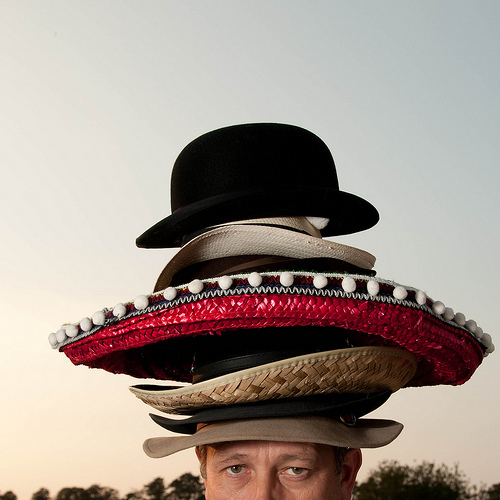}
    \includegraphics[width=0.24\linewidth]{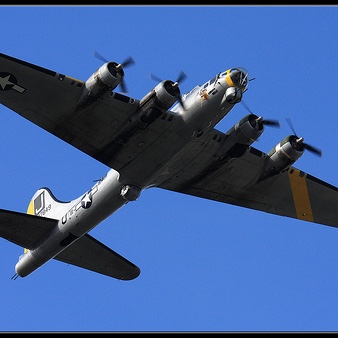}
    \caption{Example images from ImageNet used in this project.}
    \label{fig:imNet_ex}
\end{figure}

\begin{figure}
    \centering
    \includegraphics[width=0.24\linewidth]{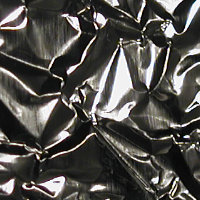}
    \includegraphics[width=0.24\linewidth]{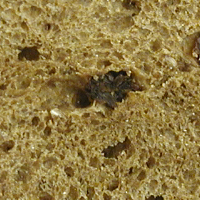}
    \includegraphics[width=0.24\linewidth]{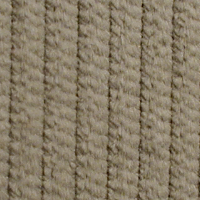}
    \includegraphics[width=0.24\linewidth]{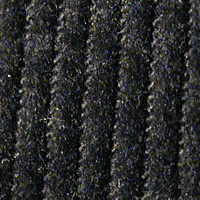}\\
    \includegraphics[width=0.24\linewidth]{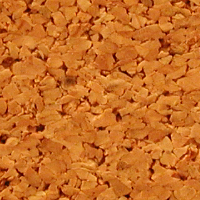}
    \includegraphics[width=0.24\linewidth]{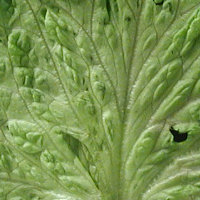}
    \includegraphics[width=0.24\linewidth]{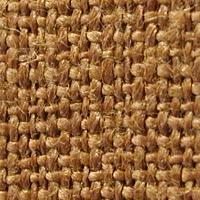}
    \includegraphics[width=0.24\linewidth]{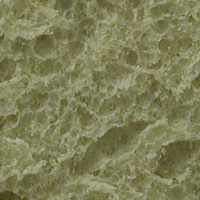}
    \caption{Example images from KTH-TIPS2 used in this project.}
    \label{fig:textures_ex}
\end{figure}

We test our method using 10,000 images from two datasets:

\begin{itemize}
    \item \textbf{Natural Images:} We use 5,000 images from the validation set of ImageNet \cite{imagenet} (see Fig. \ref{fig:imNet_ex}). Using a classification-oriented dataset gives us several images of the same class, that will be harder to differentiate for our neural system than if we just used random natural images.
    \item \textbf{Texture Images:} We use the KTH-TIPS2 dataset \cite{KTH} (see Fig. \ref{fig:textures_ex}). It contains 4754 images of 11 different textures including crumpled aluminium foil, cork, wool, lettuce leaf, corduroy, linen, cotton, brown bread, white bread, wood, and cracker. These kind of texture images would be particularly hard to remember for human participants.
\end{itemize}

\subsection{Forced-choice Test}

\begin{figure*}[t]
  \centering
   \includegraphics[width=0.9\textwidth]{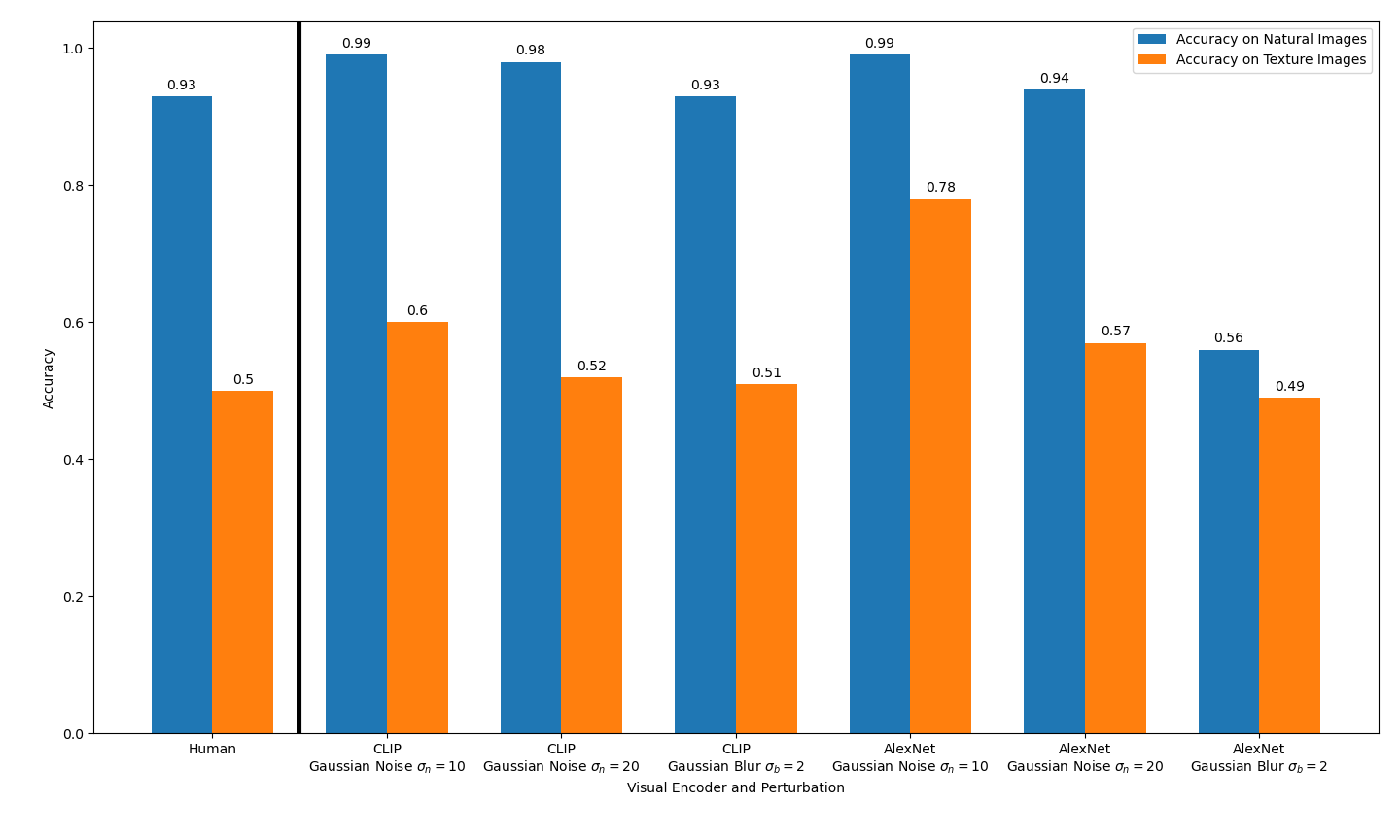}
   \caption{Forced-choice Test results with a total of 10,000 images.}
   \label{fig:forced-choice}
\end{figure*}

In this test, the neural system sees about 2,500 images, that are encoded into its memory. Then, it is presented with 2,500 forced-choices between a seen image and a novel one. The choice is made according to the method described in \ref{ssec:Remembering an image}. We test the method with both CLIP and AlexNet as encoders, as well as different levels of Gaussian noise or blur. In Fig. \ref{fig:forced-choice}, we present the results. We can see that if the noise perturbation is too low during the memory process, the system will not achieve human-like memory as accuracy on textures will be better than random. Interestingly, AlexNet achieves remarkable accuracy when the noise is low, even completely outperforming CLIP on texture images (78\% against 60\%). This seems to confirm that early convolutional classifiers were more focused on textures than on the general structure of the image. As the noise increases, the result inverts: CLIP performs better, especially on natural images. AlexNet results even drop to almost complete randomness for both natural and texture images when using blur as memory perturbation. As Gaussian blur acts as a low-pass filter, this indicates a high sensibility of AlexNet to high frequencies. With blur perturbation, CLIP maintains good accuracy on natural images (93\%) and drops to random on textures (51\%). This actually matches almost perfectly human results from \cite{oliva}. Overall, the best combination seems to be the CLIP model with a Gaussian noise of standard deviation 20, as it gives an accuracy of 98\% for natural images but stays almost random on textures (52\%). We use this model for the next task.


\subsection{Repeat-Detection Task}

The repeat-detection task is performed by streaming 2500 novel images to the neural system. About one in eight images is repeated. The system should trigger an alarm for repetitions. Results are presented in table \ref{tab:repeat-detection}. For humans, the accuracy detecting repeating images depends a lot on how long ago the image was seen for the first time \cite{oliva}. If the image is repeated immediately (1-back), the accuracy reaches 100\%. With 1023 intervening images, the accuracy drops to 80\%. For the neural system, time does not matter. Future studies could explore memory decay over time for neural systems.

\begin{table}[h]
  \centering
\begin{adjustbox}{width=1\linewidth}
  \begin{tabular}{cccc}
    \toprule
    Participant & Perturbation & Accuracy detecting repeating items & False-alarm rate \\
    \midrule
    Human & / & 80\% - 100\% & 1.0\% \\
    CLIP & Gaussian Noise $\sigma_n=20$& 89\% & 3.9\% \\
    CLIP & Gaussian Noise $\sigma_n=10$& 97\% & 2.9\% \\
    \bottomrule
  \end{tabular}
\end{adjustbox}
  \caption{Repeat-Detection task performance on natural images.}
  \label{tab:repeat-detection}
\end{table}

In table \ref{tab:repeat-detection-textures}, we present the results of the repeat-detection task on texture images. The detection of repeating images becomes almost random, and the false alarm rate dramatically increases.

\begin{table}[h]
  \centering
\begin{adjustbox}{width=1\linewidth}
  \begin{tabular}{cccc}
    \toprule
    Participant & Perturbation & Accuracy detecting repeating items & False-alarm rate \\
    \midrule
    CLIP & Gaussian Noise $\sigma_n=20$& 50\% & 41\% \\
    CLIP & Gaussian Noise $\sigma_n=10$& 56\% & 37\% \\
    \bottomrule
  \end{tabular}
\end{adjustbox}
  \caption{Repeat-Detection task performance on texture images.}
  \label{tab:repeat-detection-textures}
\end{table}

\begin{figure*}[t]
  \centering
    \subfloat[AlexNet encodings]{
      \includegraphics[width=0.45\textwidth]{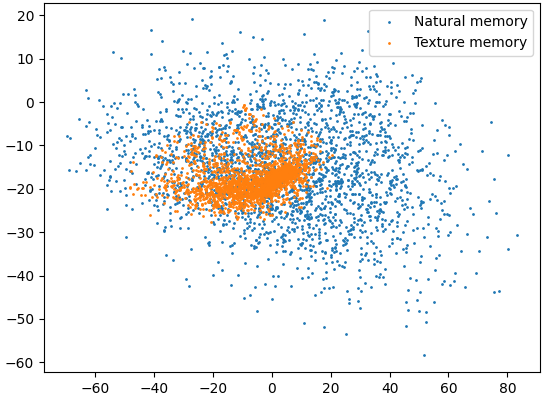}
      \label{fig:alexnet_pca}}
    \subfloat[CLIP encodings]{
      \includegraphics[width=0.45\textwidth]{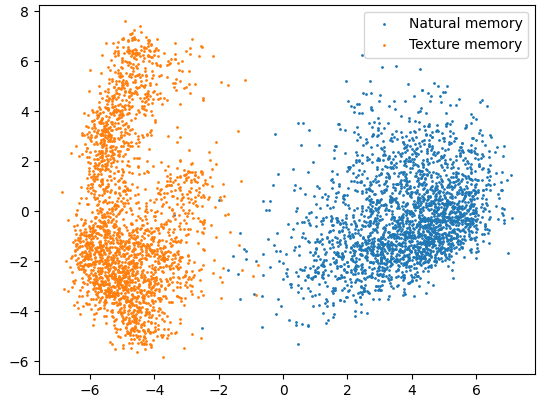}
      \label{fig:clip_pca}}
   \caption{PCA projection of the memory encodings on a plane for AlexNet (left) and CLIP (right) encodings.}
   \label{fig:pca}
\end{figure*}

\newpage
\section{Analysis}

To try to better understand the differences in performance between the two compared encoders, the memory encodings of all images were projected on a 2D plane using the Principal Component Analysis (PCA) method (see Fig. \ref{fig:pca}) \cite{pca}. Interestingly, the CLIP encoder (Fig. \ref{fig:clip_pca}) naturally separates natural images from textures. This seems to imply that the CLIP encoder looks for the high-level information mentioned in section \ref{sec:latent}, similar to how humans operate. Since such information cannot be found in textures, the encoder naturally groups them all together, thus making differentiating textures harder. On the other hand, the AlexNet encoder (Fig. \ref{fig:alexnet_pca}) seems to treat textures and natural images equally, although textures seem to be clustered slightly closer together.\\

\begin{figure}[b]
    \centering
    \includegraphics[width=\linewidth]{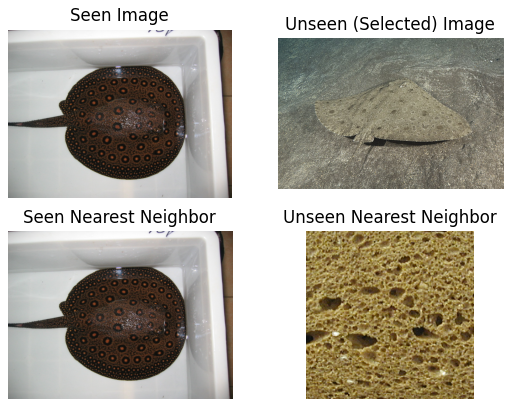}
    \caption{Failure case example for the force-choice scenario using AlexNet. The left column corresponds to the image that was already seen along with its nearest neighbor in memory; the right side corresponds to the unseen image that was mistakenly 'recognized' by the algorithm along with its nearest neighbor in the memory.}
    \label{fig:fail_alex}
\end{figure}

\begin{figure}[t]
    \centering
    \includegraphics[width=\linewidth]{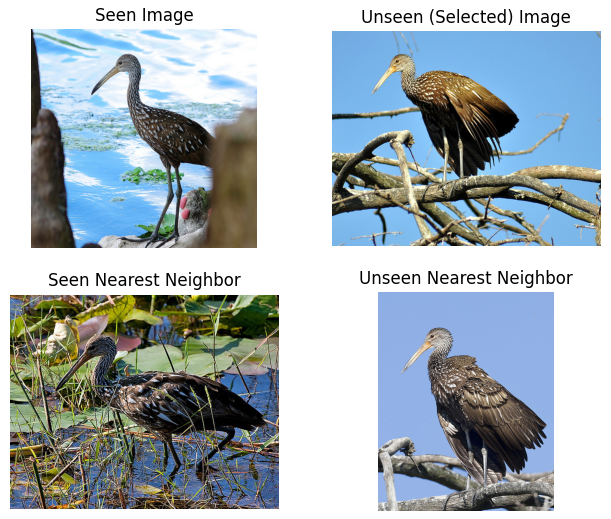}
    \caption{Failure case example for the force-choice scenario using CLIP. The left column corresponds to the image that was already seen along with its nearest neighbor in memory; the right side corresponds to the unseen image that was mistakenly 'recognized' by the algorithm along with its nearest neighbor in the memory.}
    \label{fig:fail_clip}
\end{figure}

\noindent These behaviors reflect in the observed performances, more particularly in failure cases. As shown in Fig. \ref{fig:fail_alex}, AlexNet sometimes leads to natural images being mistaken for texture images, a behavior that has never been observed with CLIP.\\
On the contrary, CLIP seems to focus more on the high-level information and actual content of the images, sometimes even slightly disregarding texture. In Fig. \ref{fig:fail_clip}, while all four images are different, all birds are the same species, face the same direction, and each pair is coherent: on the left, both pictures show a heron on water, while on the right both present a heron on a branch. In fact, even as humans, we are quite intrigued by the right pair. The colors of the two herons are slightly different (probably due to the light), their neck pose is not exactly the same and some background branches are absent in the second picture. Still, the herons are extremely similar and the bottom branches seem to match perfectly. We believe they are the same scene, taken by different cameras/angles, which makes their matching by the algorithm an interesting result.

\section{Conclusion}
\label{sec:conclusion}

Our method was able to match human performances on both repeat-detection and forced-choice memory tasks. Using pre-trained models' embedding layers (CLIP) on inputs perturbed with Gaussian noise, our algorithm reaches 97\% on natural images and 52\% on textures.
\newpage
{
    \small
    \bibliographystyle{ieeenat_fullname}
    \bibliography{main}
}


\end{document}